\pdfoutput=1

\documentclass[11pt]{article}

\usepackage[]{acl}

\usepackage{times}
\usepackage{latexsym}

\usepackage[T1]{fontenc}

\usepackage[utf8]{inputenc}

\usepackage{microtype}
\usepackage{bm}
\usepackage{booktabs} 
\usepackage{amsmath}
\usepackage{makecell}
\usepackage{amssymb}
\usepackage{enumitem}
\usepackage{multicol}
\usepackage{multirow}
\usepackage{footnote}
\usepackage{subcaption}
\usepackage{amsthm}
\usepackage{natbib}
\usepackage{diagbox}
\usepackage{caption}
\usepackage{graphicx}
\newtheorem*{hypothesis}{Hypothesis}
\newcommand{\mat}[1]{{\bf #1}}   
\DeclareMathOperator*{\argmax}{arg\,max}
\DeclareMathOperator*{\argmin}{arg\,min}
\title{Debiasing Word Embeddings with Nonlinear Geometry}

\author{Lu Cheng\\ Department of Computer Science,\\ University of Illinois Chicago \\
\textit{lucheng@uic.edu}\\ {\bf
Nayoung Kim} \and {\bf{Huan Liu}} \\School of Computing and Augmented Intelligence\\Arizona State University 
\\
\textit{\{nkim48,huanliu\}@asu.edu}\\}

\begin{document}
\maketitle
\begin{abstract}
Debiasing word embeddings has been largely limited to individual and independent social categories. However, real-world corpora typically present multiple social categories that possibly correlate or intersect with each other. For instance, ``hair weaves'' is stereotypically associated with African American females, but neither African American nor females alone. Therefore, this work studies biases associated with multiple social categories: \textit{joint biases} induced by the union of different categories and \textit{intersectional biases} that do not overlap with the biases of the constituent categories. We first empirically observe that individual biases intersect \textit{non-trivially} (i.e., over a one-dimensional subspace). Drawing from the intersectional theory in social science and the linguistic theory, we then construct an intersectional subspace to debias for multiple social categories using the nonlinear geometry of individual biases. Empirical evaluations corroborate the efficacy of our approach\footnote{Data and implementation code can be downloaded at \url{https://github.com/GitHubLuCheng/Implementation-of-JoSEC-COLING-22}.}.
\end{abstract}
\section{Introduction}
Due to the reliance on the large-scale text corpora for training, it has been observed that word embeddings are prone to express social biases inherent in the data \cite{bolukbasi2016man,caliskan2017semantics}. Prior research \citep[e.g.][] {zhao2019gender,bolukbasi2016man} in debiasing word embeddings mitigates biases associated with individual social categories and treats each category in isolation. For example, the seminal Hard-Debiasing approach \cite{bolukbasi2016man} identifies the bias direction of a category (e.g., gender) and then removes the direction from the target word such that it is equidistant to all groups (e.g., female and male at a binary level with gender) in the category. However, real-world training corpora typically present multiple social categories (e.g., gender and race), possibly with higher cardinality. These social categories can further correlate or intersect with each other \cite{thomas2004psychology,hancock2007multiplication}. Despite the promising results, debiasing for individual social categories limits our understanding of the complex nature of social biases.
\begin{figure}
    \centering
    \includegraphics[width=.7\linewidth]{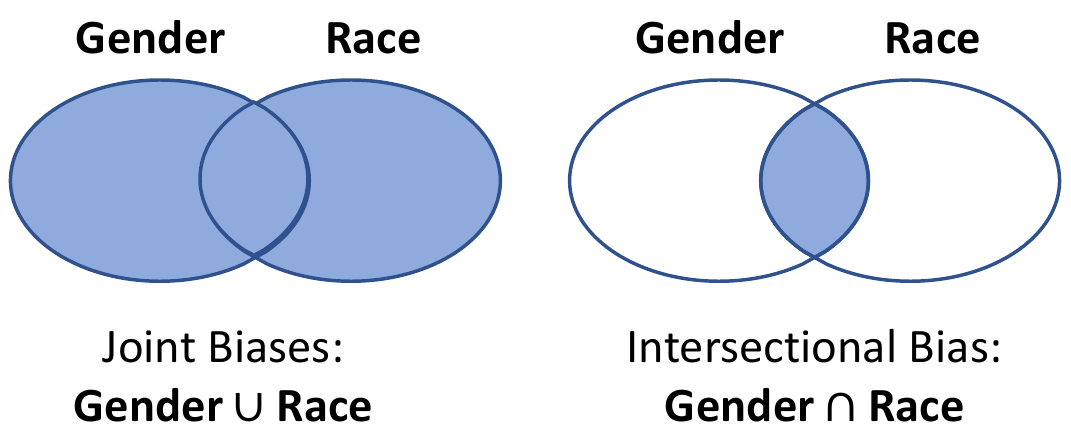}
   \caption{Illustrations of \textit{joint} (i.e., union)  and \textit{intersectional} (i.e., intersect) biases using Gender and Race.}
    \label{fig:intro}
\end{figure}

Alternatively, we might consider biases in the presence of multiple social categories \cite{foulds2020intersectional,kearns2018preventing,cheng2022bias,cheng2022toward}. This can result in at least two scenarios, as depicted in Fig. \ref{fig:intro}. First, word embeddings can simultaneously present multiple biases that might \textit{non-trivially correlate} with each other. For example, debiasing for gender can influence the results of racial bias. In the literature of social psychology, a number of works \citep[e.g.][]{akrami2011generalized,bierly1985prejudice,allport1954nature} studied the interrelationship between various biases, a.k.a. ``generalized prejudice''. We refer to bias induced by the \textbf{union} of different social categories as \textbf{joint biases}. Second, a few recent works \citep[e.g.][]{guo2020detecting} detected \textbf{intersectional biases} in word embeddings, which is the bias that does not overlap with the biases of their constituent identities. For example, ``hair weaves'' is stereotypically associated with African American females \cite{ghavami2013intersectional}.

The primary goal of this work is to mitigate the two kinds of biases in word embeddings. There are several challenges: First, it is highly possible that different biases are nonlinearly correlated \cite{cheng2022toward}. Simply taking a linear combination of individual bias subspaces (e.g., sum or mean of all bias subspaces) might lead to ineffective solutions that even amplify individual biases. Second, identifying a bias subspace typically needs predefined word sets related to target groups. These sets are curated by experts to most accurately represent each group and the associated social biases. This can be time-consuming and requires great human effort. Existing sets for identifying intersectional groups are extremely limited. They are small and exclusively used for the intersectionality of gender and race \cite{guo2020detecting,tan2019assessing}. Therefore, relying on predefined sets limits the use of debiasing approaches in practice.

To address these challenges, we first empirically observe that individual bias subspaces constructed via existing word sets intersect over a one-dimensional subspace. We then relate our findings to the intersectionality theory by \citet{crenshaw1989demarginalizing} and the linguistic theory introduced by \citet{firth1957synopsis}. The result is a \textbf{hypothesis} that the bias subspace for multiple social categories roughly resides in the intersection of all individual bias subspaces. The proposed approach (coined as \underline{Jo}int and Inter\underline{sec}tional Debiasing, JoSEC) departs from the linear correlation assumption and leverages the nonlinear geometry of subspace representations to learn an intersectional subspace. JoSEC does not need any additional human-coded defining sets for intersectional groups except for the defining sets for the constituent groups. We contribute to: 
\begin{itemize}[leftmargin=*]
    \item a novel problem that considers biases associated with multiple social categories in mitigation;
    \item an effective approach for constructing the bias subspace without relying on the defining sets for intersectional categories; and
    \item demonstrations of the effectiveness of JoSEC via empirical evaluations on the benchmark datasets.   
\end{itemize}
\section{Related Work}
Early efforts in debiasing word embeddings have been focused on gender. The seminal work by \citet{bolukbasi2016man} proposed a post-processing approach that projects gender-neutral words into a gender subspace identified by defining sets of gendered words such as \textit{she, he, man, woman}. Gender bias can then be alleviated through hard-debiasing in which the bias components in non-gendered words (e.g., \textit{doctor, nurse}) are first removed and the gendered words are then centered and equalized. \citet{manzini2019black} further extended the hard-debiasing method to multi-class settings such as race. Given a corpus, one can also learn gender-neutral word embeddings by modifying the GloVe \cite{pennington2014glove} objective function \cite{zhao2018learning}. This in-processing approach was further extended to a post-processing approach by \citet{kaneko2020autoencoding}, who suggested preserving gender-related information using autoencoder. Previous research as well as our work focuses on static word embeddings, however, stereotyped biases have also been found in contextualized word embeddings, e.g., \cite{zhao2019gender,bordia2019identifying}. 

Despite the fruitful results, most prior works were found to remove biases superficially and fail to deliver gender-neutral embeddings \cite{gonen2019,blodgett2020language}. Therefore, it is increasingly observable that existing bias removal techniques are insufficient to guarantee gender-neural modeling. The majority of existing works were also criticized for not examining the impact of gender bias in real-world applications \cite{blodgett2020language}. In contrast to prior research focused on one form of bias in debiasing word embeddings, this work aims to provide a simple yet effective approach for bias mitigation in the presence of multiple bias forms. As human-like biases exist in the majority of word embeddings and debiasing approaches are unlikely to largely affect our results, we build our approach upon the seminal Hard-Debiasing algorithm by \citet{bolukbasi2016man}. Future research is warranted to investigate other debiasing approaches.

For intersectional bias, most of the existing works are in social science and psychology literature, such as \cite{crenshaw1989demarginalizing,kahn1989psychology,hare1988meaning}. Comparatively fewer efforts can be found in the computer science field. One such work \cite{buolamwini2018gender} examined the intersectional accuracy disparities in commercial gender classification systems. For contextualized word embeddings, \citet{may2019measuring} and \citet{tan2019assessing} measured the emergent intersectional biases of African American females using attributes presented in \cite{caliskan2017semantics}. In complement to WEAT, \citet{guo2020detecting} proposed the Contextualized Embedding Association Test (CEAT) to measure the intersectional bias. These methods detect intersectional bias using the defining and attribute sets related to intersectional groups.

In summary, this work complements prior research by mitigating biases related to multiple social categories. JoSEC does not rely on the human-coded word sets used to define the intersectional groups to identify the bias subspace. While this work focuses on non-contextualized embeddings, research \citep[e.g.,][]{guo2020detecting,lepori2020unequal} has found contextualized word embeddings such as BERT \cite{devlin2019bert} display intersectional biases. We leave it for future exploration.
\section{Preliminary}
In this section, we briefly review the backbone model of this work: the seminal hard-debiasing method \cite{bolukbasi2016man} and its extension. It consists of two steps: identifying the bias subspace and removing bias components.
\subsection{Identifying Bias Subspace}
The individual bias subspace is identified by the \textit{defining sets}, in which words represent different ends of the bias. For example, the defining sets of gender can be the gendered pronouns \textit{\{he, she\}} and nouns \textit{\{man, woman\}}. One can then identify the gender subspace $\mathcal{B}$ by (1) subtracting the word embeddings of words in each defining set from the set's mean, and (2) obtaining the $K$ most significant components of the resulting vectors through a dimensionality-reduction method.
\subsection{Removing Bias Components}
\label{sec:removing}
The next step is to apply the hard debiasing strategy to completely or partially remove the subspace components from the word embeddings. Hard-debiasing consists of two steps -- \textit{Neutralize} and \textit{Equalize}. ``Neutralize'' removes bias components from non-gendered words (e.g., \textit{doctor} and \textit{nurse}); ``Equalize'' aims to center the gendered word embeddings (e.g., \textit{she} and \textit{he}) and equalize their bias components, such as the word pair \textit{\{man, woman\}}.

Formally, given a bias subspace $\mathcal{B}=\{\bm{b_1},\bm{b_2},...,\bm{b_K}\}$, where $K$ denotes the number of principal components, we first compute the bias component $\mat{w}_{\mathcal{B}}$ of embedding $\mat{w}\in\mathbb{R}^d$ in $\mathcal{B}$ by
\begin{equation}
 \small
    \mat{w}_{\mathcal{B}}=\sum_{k=1}^K\big<\mat{w},\bm{b_k}\big>\bm{b_k}.
    \label{subspace}
\end{equation}
We then \textit{neutralize} word embeddings by removing the bias component from non-gendered words: 
\begin{equation}
 \small
    \mat{w}'=\frac{\mat{w}-\mat{w}_{\mathcal{B}}}{\|\mat{w}-\mat{w}_{\mathcal{B}}\|},
    \label{neutralize}
\end{equation}
where $\mat{w}'$ are the debiased word embeddings.

To ``Equalize'', we debias the gendered words in a given equality set $E$ by the following equation:
\begin{equation}
\small
    \mat{w}'=(\bm{\mu}-\bm{\mu}_{\mathcal{B}})+\sqrt{1-\|\bm{\mu}-\bm{\mu}_{\mathcal{B}}\|^2}\frac{\mat{w}_{\mathcal{B}}-\bm{\mu}_{\mathcal{B}}}{\|\mat{w}_{\mathcal{B}}-\bm{\mu}_{\mathcal{B}}\|},
    \label{equalize}
\end{equation}
where $\bm{\mu}=\frac{1}{|E|}\sum_{\mat{w}\in E}\mat{w}$ is the mean of embeddings of the words in the equality set $E$. $\bm{\mu}_{\mathcal{B}}$ denotes the bias component of $\bm{\mu}$ in the identified bias subspace. It can be obtained via Eq. \ref{subspace}. 
\subsection{Extending into Multi-Class Settings}
In a multi-class setting (i.e., with more than two classes in a category, e.g., religion or race), the task inherently becomes non-linearly separable  \cite{manzini2019black}. However, it is possible to linearly separate multiple classes based on the components of word embeddings. The multi-class bias subspace is then defined as follows: Given $n$ defining sets of word embeddings $\{D_{1},D_{2},...,D_{n}\}$, the bias subspace $\mathcal{B}$ is defined by the first $K$ components of the following Principal Component Analysis (PCA) \cite{abdi2010principal} evaluation:
\begin{equation}
\small
    \mathcal{B}=\textbf{PCA}\Big(\bigcup_{i=1}^{n}\bigcup_{\mat{w}\in D_{i}}\mat{w}-\bm{\mu}_{i}\Big),
\end{equation}
where $\bm{\mu}_{i}=\frac{1}{|D_{i}|}\sum_{\mat{w}\in D_{i}}\mat{w}$ is the mean of word embeddings in set $i$. $\bigcup$ denotes concatenation by rows. To remove multi-class bias, one can use the hard-debiasing method described in Sec. \ref{sec:removing}.
\section{Method}
Existing approaches for debiasing word embeddings work on individual categories, rendering incomplete measurement of various social biases \cite{hancock2007multiplication,hurtado2008more}. To account for biases associated with multiple social categories, we need to address the primary challenges of the potential non-linear correlations between biases and the difficulty of curating defining sets to identify the bias subspace for these categories. In this section, we introduce the proposed approach -- JoSEC -- for identifying such a subspace \textbf{without additional} human-coded defining set, which we refer to as the \textit{intersectional subspace}.
\subsection{Social Categories as ``Cultural Contexts''}
If individual biases are linearly correlated, we may construct the intersectional subspace by simply taking the sum or average of all individual bias subspaces. However, social biases are complex by nature as suggested by evidence in social science and psychology such as the ``generalized prejudice'', i.e., generalized devaluing sentiments across different groups. To better quantify the potential non-linearity, we might first take a step back and revisit the development of single-word embeddings. 

An influential position in the development of word embeddings holds that semantic representations for words can be derived through the patterns of lexical co-occurrence in language corpora. This is famously summarized by \citet{firth1957synopsis} as ``you shall know a word by the company it keeps''. The central tenet is the idea that the sense of the target word could be inferred from its contexts, i.e., neighboring words within the sentence. Informed by this finding, we might assume that human-like biases follow a similar principle: a bias form w.r.t. some social category can be identified by its unique cultural contexts. In debiasing word embeddings, this indicates that the bias subspace (e.g., gender subspace) can be defined by the defining sets of words (e.g., \textit{she, he}) that provide a specific ``cultural context'' for this bias form. 

Naturally, when defining the intersectional subspace associated with multiple social categories, \textit{we might similarly consider each social category as a unique ``cultural context'' of the intersectional subspace}. For example, the subspace of the intersectional group of gender and race is defined by both the gender subspace and the race subspace. That is, ``Aisha'' -- a common name of an African American female -- can be exclusively defined within a cultural context jointly determined by both gender and race whilst terms such as ``hair weaves'' should not depend on such context. Underpinning this assumption is the idea similar to the linguistic theory: each social category provides \textit{unique context} to construct the intersectional subspace associated with multiple categories.  
\subsection{Geometry of Subspace Representation}
Under the ``social categories as cultural contexts'' assumption, the intersectional subspace might have a fairly large intersection with each individual bias subspace. We empirically observe and hypothesize that the intersectional subspace of multiple social categories should reside in all the subspaces representing the ``cultural contexts'' where the intersectional subspace is defined. Specifically, the intersectional subspace resides in the \textit{intersection} of all individual bias subspaces and these subspaces should \textbf{intersect non-trivially}. This further implies that there exists a direction (one-dimension subspace) that is extremely close to all individual bias subspaces. We use this vector to represent the intersectional subspace. We propose the following hypothesis for identifying the bias subspace associated with multiple social categories:
\begin{hypothesis}[\textit{Intersectional Hypothesis}] The intersectional subspace $\mathcal{B}_{sec}$ should reside in the intersection of $\{\mathcal{B}_1, \mathcal{B}_2,... \mathcal{B}_N\}$, where $\mathcal{B}_i$ denotes the bias subspace of social category $i$ and $N$ is the number of considered social categories.
\end{hypothesis}
Intersectional Hypothesis can be seen as operationalizing the intersectionality theory guided by Firth's hypothesis. 
\begin{figure}
    \centering
    \includegraphics[width=.65\linewidth]{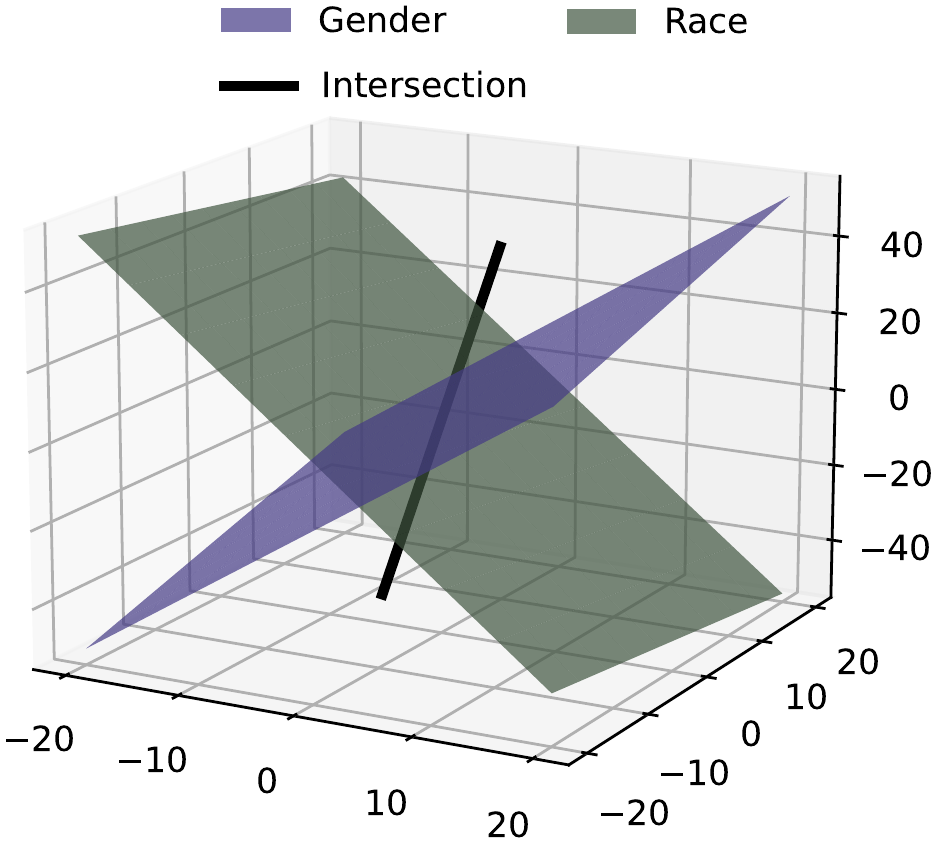}
   \caption{The geometry of Gender and Race bias subspaces and the intersectional subspace (Intersection).}
\label{fig:visualize}
\end{figure}
\begin{figure}
    \centering
    \includegraphics[width=.9\linewidth]{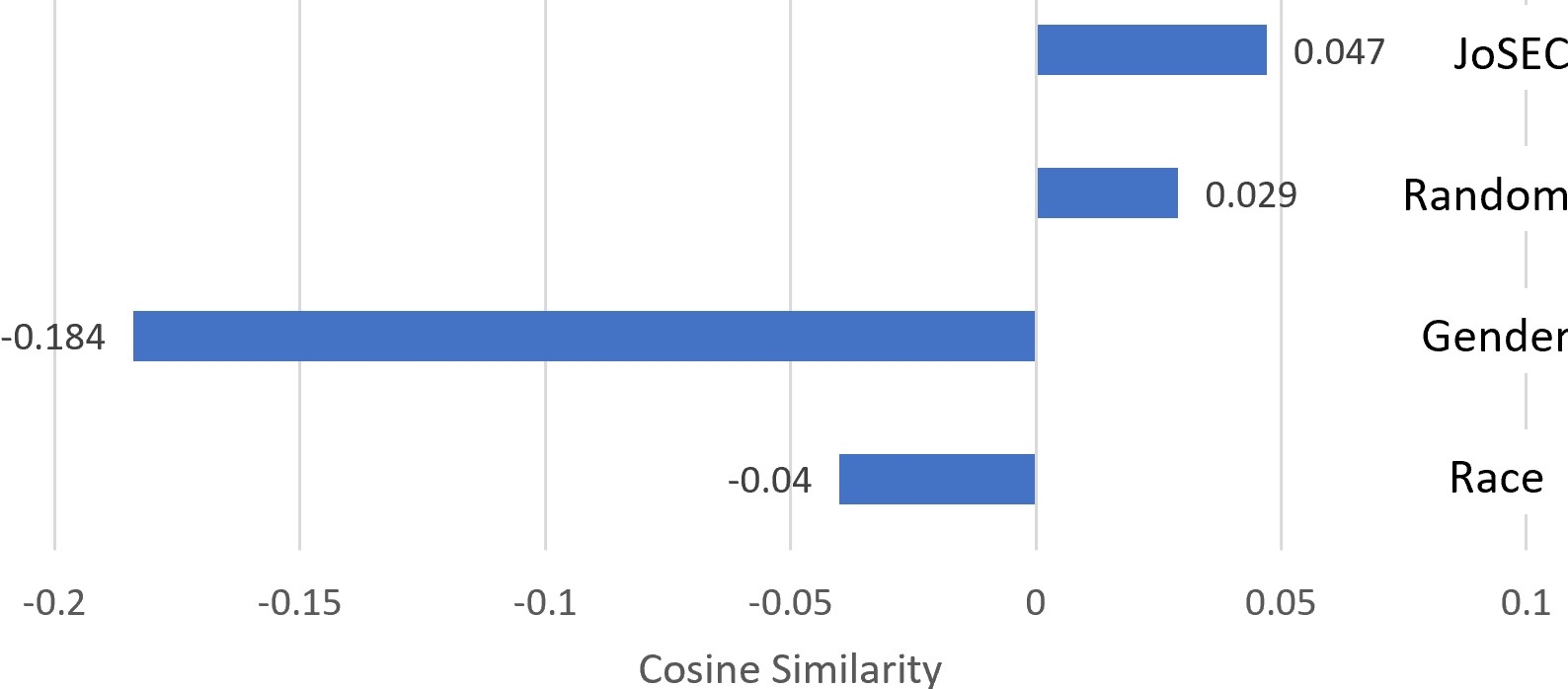}
   \caption{Cosine similarity between the ground-truth intersectional subspace and subspaces obtained via various approaches. ``Random'' denotes the result averaged over 10 similarity scores from 10 random vectors.}
    \label{cossim}
\end{figure}
\subsubsection{Empirical Validation of the Intersectional Hypothesis}
We empirically validate the intersectional hypothesis using the benchmark dataset L2-Reddit corpus (detailed in Sec. \ref{data}). The dataset includes the defining sets for three social categories: race, gender, and religion, respectively. We first construct three individual bias subspaces for the target categories and each subspace has dimensions $K\times d$, where $d$ denotes the dimension of word embeddings. For better visualization, we further project the $d$-dimensional subspace representations to 3D vectors using PCA.

We draw subspaces w.r.t. two randomly selected categories as 2-dimensional planes in Fig. \ref{fig:visualize}. Result for the three categories is in Appendix A. We also visualize the corresponding intersectional subspaces identified by Eq. \ref{sim_intersect}. We observe that the individual bias subspaces intersect roughly in a common direction with which the intersectional subspace approximately aligns. In addition, we use the defining sets for the intersectional groups of gender and race (e.g., African American Female) provided by WEAT \cite{caliskan2017semantics} and \citet{parada2016ethnolinguistic} to construct the ``ground-truth'' intersectional bias subspace. We then calculate the cosine similarity between the ground-truth subspace and the (a) gender subspace, (b) race subspace, (c) random vector, and (d) intersectional subspace approximated by JoSEC, respectively. The similarity score ranges from -1 to 1, with -1 denoting the most dissimilar. Results in Fig. \ref{cossim} suggest that JoSEC generates the intersectional subspace significantly more similar to the ground-truth subspace. Note that most of the similarity scores are close to 0. We believe this is in part because of the limitation of existing defining sets for intersectional identities, e.g., small in size. Both quantitative (Fig. \ref{cossim}) and qualitative (Fig. \ref{fig:visualize}) analyses empirically justify the intersectional hypothesis.  
\subsubsection{Identifying Intersectional Subspace} Under the \textit{Intersectional Hypothesis}, we are essentially seeking the direction vector $\mathcal{B}_{sec}=\hat{\bm{u}}$ that is ``closest'' to all individual bias subspaces. Let $\bm{u}$ be a unit-length vector. We then reduce the problem of identifying intersectional subspace to the following optimization task:
\begin{equation}
\small
    \hat{\bm{u}}=\argmin_{\|\bm{u}\|=1}\sum_{i=1}^N d(\bm{u}, \mathcal{B}_i)^2,
    \label{intersect}
\end{equation}
where $d(\bm{u}, \mathcal{B}_i)$ is the shortest $\ell_2$-distance between the intersectional subspace and the bias subspace of social category $i\in\{1,2,...,N\}$. Formally, 
\begin{equation}
\small
    d(\bm{u}, \mathcal{B}_i)=\sqrt{\|\bm{u}\|^2-\sum_{k=1}^K(\bm{u}^\intercal \bm{v}_{ik})^2},
\end{equation}
where $\{\bm{v}_{i1}, ..., \bm{v}_{iK}\}$ are the $K$ principal components representing bias subspace $\mathcal{B}_i$. Eq. \ref{intersect} can be reformulated as the following:
\begin{equation}
\small
        \hat{\bm{u}}=\argmax_{\|\bm{u}\|=1}\sum_{i=1}^N\sum_{k=1}^K (\bm{u}^\intercal\bm{v}_{ik})^2.
        \label{sim_intersect}
\end{equation}
Eq. \ref{sim_intersect} can be solved by taking the first principal component of $\{\bm{v}_{ik}\}_{i=1,...,N;k=1,...,K}$.

With the identified intersectional subspace $\hat{\bm{u}}$, we then follow Eq. \ref{subspace}-\ref{equalize} to remove the identified intersectional bias components from the target words (e.g., \textit{hair weaves}).
\section{Experiments}
We validate the efficacy of the proposed intersectional subspace for debiasing word embeddings with multiple social categories. We answer the following research questions: How does JoSEC fare against baselines for mitigating (\textbf{RQ. 1}) \textit{joint} biases and (\textbf{RQ. 2}) \textit{intersectional} biases? How does JoSEC influence the (\textbf{RQ. 3}) utility of word embeddings in downstream tasks and  
(\textbf{RQ. 4}) biases in downstream tasks such as toxicity detection?
\subsection{Language Corpus and Social Bias Data}
\label{data}
Results for debiasing word embeddings are based on the commonly-used L2-Reddit corpus \cite{rabinovich2018native}, a collection of Reddit\footnote{https://www.reddit.com/} posts and comments. It has been shown that the structural factor in user-generated content sites like Reddit make them less welcoming to marginalized populations \cite{bender2021dangers}. The initial biased word embeddings are obtained by training word2vec on approximately 56 million sentences. It includes three social categories: \textit{Gender, Race}, and \textit{Religion}. We use vocabularies from \cite{bolukbasi2016man} and \cite{caliskan2017semantics} as the defining and attribute sets for gender. Word sets for race and religion are the same lexicons used in \cite{manzini2019black}. 

Following \cite{guo2020detecting}, we consider the intersectionality of race and gender for the evaluation of intersectional debiasing due to the limited data availability. In particular, there are in total $3\times2$ (3 racial classes and 2 gender classes) intersectional social groups: \{\textit{African American male, African American female, European American male, European American female, Mexican American male, Mexican American female}\}. The defining sets for these groups are provided by WEAT \cite{caliskan2017semantics} and \citet{parada2016ethnolinguistic}, including frequent given names that represent group membership. The intersectional attribute sets identified through human workers' validation are provided by \citet{ghavami2013intersectional}. For example, one common given name included in the defining set of African American females is ``Aisha'' and some related stereotyped bias include attributes ``aggressive'' and ``dark-skinned''. For the complete list of given names and attributes, please refer to Appendix B. Note that in contrast to common debiasing approaches, JoSEC does not need these defining sets for joint and intersectional debiasing. We use them for evaluation only.
\subsection{Experimental Setup}
We briefly summarize the experimental settings, including the baselines, downstream tasks, and evaluation metrics.
\subsubsection{Baselines}
\label{baseline}
We are not aware of any existing approaches for joint or intersectional debiasing for word embeddings. Therefore, we adapt the hard-debiasing approach such that it works in both tasks. We also consider two debiasing strategies (i.e., SUM and MEAN) that impose a linear assumption to construct the bias subspace for multiple social categories. All compared approaches only differ in the subspace construction, the bias mitigation follows the same procedure in the hard-debiasing approach. Baselines are detailed below.
\begin{itemize}[leftmargin=*]
    \item \textit{Hard\_Seq}. The hard-debiasing method extended to the \textit{joint} debiasing task. In particular, \textit{Hard\_Seq} sequentially debiases for individual social categories. As the order might influence the results of \textit{Hard\_Seq}, we experiment with all potential sequences and report the best results.
    \item \textit{Hard\_Insec}. The hard-debiasing method extended to the \textit{intersectional} debiasing task. In particular, \textit{Hard\_Insec} uses the human-coded defining sets for the intersectionality groups of gender and race in \cite{guo2020detecting} to construct the intersectional subspace. 
    \item \textit{SUM}. Its bias subspace is constructed by summing up the subspaces of individual biases, i.e., $\mathcal{B}_{SUM}=\sum_{i=1}^N\mathcal{B}_i.$

    \item \textit{MEAN}. Its bias subspace is constructed by averaging over the subspaces of individual biases, i.e.,
        $\mathcal{B}_{MEAN}=\frac{1}{N}\sum_{i=1}^N\mathcal{B}_i.$

\end{itemize}
\subsubsection{Downstream Tasks} 
\textbf{Utility (RQ. 3): }To examine the influence of JoSEC on the utility of word embeddings, we perform several standard downstream tasks following \cite{manzini2019black}. They are the CoNLL 2003 shared tasks \cite{sang2003introduction}, including NER tagging, POS (part-of-speech) tagging, and POS chunking. There are two evaluation paradigms: replacing the biased embeddings with the debiased ones or retraining the model on debiased embeddings. We only report results for one setting and the other can be found in Appendix C.

\noindent\textbf{Extrinsic Bias (RQ. 4):} While we work on mitigating biases in the pre-trained resource, i.e., \textit{intrinsic bias}, recent research \citep[e.g.,][]{seraphina2021intrinsic,delobelle2021measuring} presents interesting findings about the biases in downstream tasks enabled by word embeddings, i.e., \textit{extrinsic bias}. Therefore, we further investigate how the debiased word embeddings influence biases in a common downstream NLP task, toxicity detection. Particularly, we consider the Kaggle Challenge of Jigsaw Unintended Bias in Toxicity Classification\footnote{https://www.kaggle.com/c/jigsaw-unintended-bias-in-toxicity-classification} and examine biases against gender, race, and religion. The Perspective API’s Jigsaw dataset has both toxicity and identity annotations. The training and test splits are the same as the original data. Please refer to Appendix D for detailed experimental settings for toxicity classification. 
\subsubsection{Evaluation Metrics}
For debiasing tasks, we use the mean average cosine similarity (MAC) to quantify the intrinsic bias, as suggested in \cite{manzini2019black}. Given a set of target word embeddings $\mathcal{S}$ of words with a specific form of social bias (e.g., \textit{Jew, Christian, Muslim}) and a set of attribute sets $\mathcal{A}=\{A_1, A_2,...,A_N\}$, $A_j$ consists of embeddings of words $\bm{a}$ (e.g., \textit{violent, terrorist, uneducated}) that should not be associated with any word in $\mathcal{S}$. Let $f(\cdot)$ be a function that computes the mean cosine distance between $S_i\in \mathcal{S}$ and $\bm{a}\in A_j$:
\begin{equation}
\small
    f(S_i,A_j)=\frac{1}{|A_j|}\sum_{\bm{a}\in A_j}\cos(S_i,\bm{a}),
\end{equation}
where $\cos(S_i,\bm{a})=1-\frac{S_i\cdot\bm{a}}{\|S_i\|_2\cdot\|\bm{a}\|_2}.$
MAC is then computed by
\begin{equation}
\small
    \text{MAC}(\mathcal{S},\mathcal{A})=\frac{1}{|\mathcal{S}||\mathcal{A}|}\sum_{S_i\in \mathcal{S}}\sum_{A_j\in \mathcal{A}}f(S_i,A_j).
    \label{mac}
\end{equation}
A larger MAC score denotes a greater bias removal.

For the downstream task that examines the utility, we report F1 scores from using biased word embeddings as well as the changes of F1 ($\Delta$ F1), Precision ($\Delta$ Precision), and Recall ($\Delta$ Recall) after using debiased word embeddings. To check the statistical significance, we also perform a paired $t$-test on the distribution of average cosine distance used to compute MAC and student $t$-test for the results of downstream tasks. For the downstream task that examines the extrinsic bias, we use two common evaluation metrics: False Positive Equality Difference (FPED) and False Negative Equality Difference (FNED). We report Total (FPED+FNED) scores from using biased word embeddings and $\Delta$ Total, $\Delta$ FPED, and $\Delta$ FNED after using debiased word embeddings. Unless otherwise noted, all the results below are statistically significant at level 0.05.
\begin{table}[]
\caption{MACs ($\uparrow$) of all approaches for joint debiasing. ``Re'', ``Ra'', and ``Ge'' denotes ``Religion'', ``Race'', and ``Gender'', respectively. ``Re$\rightarrow$Ra$\rightarrow$Ge'' denotes the order for sequential debiasing. We experiment with all potential sequences and report the best results. A larger value is more desired.}
\resizebox{\columnwidth}{!}{\begin{tabular}{|c|c|c|c|c|}
\hline
        \diagbox{Model}{MAC}                                                       & Gender         & Race           & Religion       & Total          \\ \hline\hline
Biased                                                         & 0.623          & 0.892          & 0.859          & 2.374         \\ \hline
\begin{tabular}[c]{@{}c@{}}Hard\_Seq\\ (Re$\rightarrow$Ra$\rightarrow$Ge)\end{tabular} & 0.656          & 0.888          & \textbf{0.937} & 2.481          \\ \hline
\begin{tabular}[c]{@{}c@{}}Hard\_Seq\\ (Ra$\rightarrow$Re$\rightarrow$Ge)\end{tabular} & 0.654          & \textbf{0.929} & 0.868          & 2.451          \\ \hline
SUM                                                            & 0.598          & 0.870          & 0.900          & 2.368         \\ \hline
MEAN                                                           & 0.657          & 0.872          & 0.862          & 2.391          \\ \hline
JoSEC                                                          & \textbf{0.703} & 0.914          & 0.917          & \textbf{2.534} \\ \hline
\end{tabular}}
\label{tab:joint}
\end{table}
\subsection{Results}
We first present results (averaged over 5 repetitions) for \textbf{RQ. 1-4} and then discuss our findings. 
\subsubsection{RQ. 1: Joint Debiasing}
The joint debiasing task seeks to simultaneously mitigate biases induced by the union of all social categories. We report MACs w.r.t. each individual bias as well as the total bias, which is computed over all considered social categories. All the best results are highlighted. 

\begin{figure}
    \centering
    \includegraphics[width=.75\linewidth]{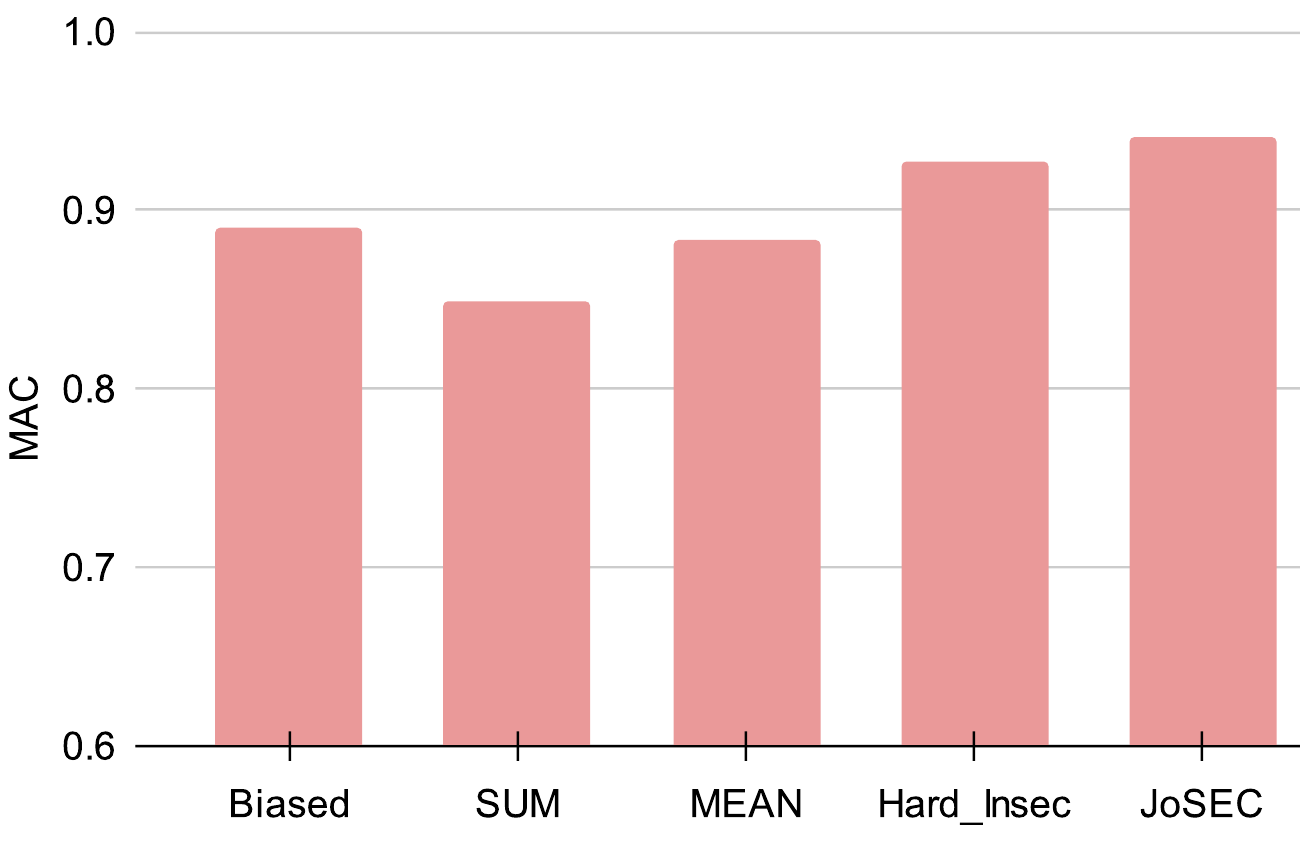}
    \caption{MACs ($\uparrow$) of all compared approaches for the intersectional debiasing task. A larger value is more desired.}
    \label{fig:inter}
\end{figure}
From the results in Table \ref{tab:joint}, we can observe that (1) the best results w.r.t. debiasing for individual categories are achieved by various approaches and the proposed approach (\textit{JoSEC}) outperforms all baselines regarding reducing the total amount of bias (Total). This suggests that it is challenging to debias for all social categories simultaneously and the proposed intersectional subspace is effective for joint debiasing. (2) The linear solutions to subspace construction  (i.e., \textit{SUM} and \textit{MEAN}) are not as effective as sequential debiasing. This empirically validates our hypothesis that social biases are non-linearly correlated and multiple social categories should intersect non-trivially.

(3) Of particular interest is that when debiasing sequentially, the bias mitigation performance w.r.t. the first category appears to be the most effective. For example, \textit{Hard\_Seq} (Re$\rightarrow$Ra$\rightarrow$Ge) shows the best MAC of ``Religion''. Further, by comparing the results in the second row (\textit{Biased}) with those in the third (\textit{Hard\_Seq}), the racial MAC of sequential debiasing decreases (i.e., more biases) whilst applying hard-debiasing to racial bias alone can actually lead to higher MACs. These findings might suggest that different biases are \textit{interacting} with each other, i.e., they are correlated. Future research is warranted to examine the bias correlations and their influence on the debiasing approaches. 
\begin{table*}[t]
\centering
\caption{Utility of biased and debiased word embeddings in NER Tagging (NER), POS Tagging (POS-T), and POS Chunking (POS-C) tasks, under the \textit{Embedding Matrix Replacement} paradigm. Word embeddings are debiased by hard-debiasing with different subspaces. $\Delta$ denotes the change before and after debiasing.}
\resizebox{\textwidth}{!}{%
\begin{tabular}{|c|c|c|c|c|c|c|c|c|c|c|c|c|c|c|c|}
\hline
                   & \multicolumn{3}{c|}{Hard\_Seq} & \multicolumn{3}{c|}{Hard\_Insec} & \multicolumn{3}{c|}{SUM} & \multicolumn{3}{c|}{MEAN} & \multicolumn{3}{c|}{JoSEC} \\ \hline
Tasks              & NER      & POS-T    & POS-C   & NER       & POS-T    & POS-C    & NER    & POS-T  & POS-C  & NER     & POS-T  & POS-C  & NER     & POS-T   & POS-C  \\ \hline
Biased F1          & 0.96     & 0.99     & 1.00    & 0.96      & 0.99     & 1.00     & 0.96   & 0.99   & 1.00   & 0.96    & 0.99   & 1.00   & 0.96    & 0.99    & 1.00   \\ \hline
$\Delta$ F1        & -0.01    & 0.01     & 0.00    & -0.01     & 0.00     & 0.00     & -0.01  & 0.01   & 0.00   & -0.01   & 0.00   & 0.00   & -0.02   & 0.00    & 0.00   \\ \hline
$\Delta$ Precision & -0.01    & 0.00     & 0.00    & -0.02     & 0.00     & 0.00     & -0.02  & 0.00   & 0.00   & -0.02   & 0.00   & 0.00   & -0.03   & 0.00    & 0.00   \\ \hline
$\Delta$ Recall    & -0.01    & 0.03     & 0.02    & -0.02     & 0.01     & 0.02     & -0.02  & 0.02   & 0.01   & -0.02   & 0.01   & 0.01   & -0.03   & 0.01    & 0.01   \\ \hline
\end{tabular}%
}
\label{downstream1}
\end{table*}
\begin{table*}[t]
\centering
\caption{Gender, racial, and religious biases in toxicity classification using biased and debiased word embeddings. Total=FPED+FNED ($\downarrow$). A smaller bias score denotes less bias. A negative $\Delta$ indicates reduced bias.}
\resizebox{\textwidth}{!}{%
\begin{tabular}{|c|c|c|c|c|c|c|c|c|c|c|c|c|c|c|c|}
\hline
                   & \multicolumn{3}{c|}{Hard\_Seq} & \multicolumn{3}{c|}{Hard\_Insec} & \multicolumn{3}{c|}{SUM} & \multicolumn{3}{c|}{MEAN} & \multicolumn{3}{c|}{JoSEC} \\ \hline
Bias              & Gender      & Race    & Religion   & Gender      & Race    & Religion    & Gender      & Race    & Religion  & Gender      & Race    & Religion  & Gender      & Race    & Religion \\ \hline
Biased Total          & 1.27     & 0.64     & 0.37    & 1.27     & 0.64     & 0.37     & 1.27     & 0.64     & 0.37   & 1.27     & 0.64     & 0.37   & 1.27     & 0.64     & 0.37  \\ \hline
$\Delta$ Total        & -0.04  & 0.04     & 0.05    & -0.05     & -0.01     & 0.01     & 0.00  & 0.04   & -0.01   & 0.00   & -0.01   & 0.01   & -0.05   & 0.02    & 0.02   \\ \hline
$\Delta$ FPED & -0.09    & 0.03     & 0.00    & -0.05     & 0.03     & -0.02     & -0.03  & 0.03   & -0.01   & 0.06   & 0.04   & 0.01   & -0.04   & 0.03    & 0.00   \\ \hline
$\Delta$ FNED    & 0.05    & 0.01     & 0.05    & 0.00     & -0.04     & 0.02     & 0.03  & 0.01   & 0.01   & -0.06   & -0.04   & 0.00   & 0.00   & -0.01    & 0.02   \\ \hline
\end{tabular}%
}
\label{toxicity}
\end{table*}
\subsubsection{RQ. 2: Intersectional Debiasing}
To evaluate the intersectional bias removal performance of all the compared methods, we use the human-coded defining and attribute sets associated with the intersectionality of gender and race. Note that only \textit{Hard\_Insec} used the human-coded defining sets to construct the bias subspace. All other approaches (i.e., \textit{SUM}, \textit{MEAN}, and \textit{JoSEC}) use subspaces of individual biases to construct the subspace for multiple social categories. MACs of all methods are presented in Fig. \ref{fig:inter}.

We make the following observations: (1) \textit{JoSEC} is most effective for mitigating the intersectional biases w.r.t. gender and race. The fact that it outperforms \textit{Hard\_Insec} manifests the potential to leverage the nonlinear geometry of subspace representation to construct the subspace for intersectional bias. This is encouraging as it suggests that we may not need defining sets for intersectional groups, which are often inaccessible  and challenging to collect; (2) \textit{SUM} and \textit{MEAN} are outperformed by \textit{Biased}, indicating that simply taking a linear combination of all the individual bias subspaces can aggravate the intersectional biases in word embeddings. The violation of the linear correlation assumption has a negative influence on the performance of hard-debiasing. 
\subsubsection{RQ. 3: Downstream Utility}
\label{downstream}
This research question aims to investigate the effects of various debiasing strategies on the semantic utility of word embeddings in standard NLP tasks. We consider NER Tagging, POS Tagging, and POS Chunking, following \cite{manzini2019black}.

Results for embedding matrix replacement are in Table \ref{downstream1}. We can observe that debiasing word embeddings with multiple social categories only slightly changes the semantic utility. We also perform the student $t$ test and further testify that these differences are statistically insignificant. These results imply that debiasing for multiple categories using the hard-debiasing method does not have a significant influence on the semantic utility of word embeddings. This applies to both joint and intersectional debiasing. 
\subsubsection{RQ. 4: Bias in Toxicity Classification}
This experiment further examines the extrinsic biases of word embeddings, particularly, in toxicity classification. Results are shown in Table \ref{toxicity}. We observe that the debiased word embeddings have little influence on reducing biases in the downstream task, indicating no reliable correlation between the \textit{intrinsic bias} and \textit{extrinsic bias}. This aligns well with findings shown in \cite{seraphina2021intrinsic}. Aware of the importance of mitigating biases in downstream applications, it is critical to extend this work to joint and intersectional debiasing focused on extrinsic measures of biases in the future.
\section{Discussions}
This work studies \textbf{joint biases} induced by the union of multiple categories and the \textbf{intersectional biases} that do not overlap with biases of the constituent categories. Challenges arise from the potential nonlinearity between different biases and the difficulty of curating human-coded word lists for identifying intersectional bias subspace. We first empirically showed that different biases intersect \textit{non-trivially}. Informed by the intersectionality theory and the linguistic theory by Firth, we propose a simple yet effective approach (JoSEC) for constructing the intersectional bias subspace using the nonlinear geometry of bias subspaces. JoSEC can reduce intrinsic bias without losing the semantic utility of word embeddings. The broad result of this research is that it is critical to consider biases associated with multiple social categories given the complex nature of human-like biases.

We do note several limitations of this study. First, the empirical observation of the nonlinearity between different biases needs more rigorous theoretical proof/evidence. A potential result would be an in-depth analysis of bias correlations in a variety of NLP tasks. Questions such as ``how are biases correlated? Negative or positive correlation?'' might be investigated. Second, research should be conducted to study how to reduce joint and intersectional biases in contextualized word embeddings such as BERT and GPT-3. Third, JoSEC focuses on intrinsic measures, which remain good descriptive metrics for computational social science \cite{seraphina2021intrinsic}. However, it might not be relied on to mitigate biases in downstream applications. As real-world scenarios can be more challenging and complicated (e.g., biases induced in the deployment), it is necessary to extend this research to extrinsic measures of biases, which might vary in different applications. Meanwhile, research needs to focus on a more rigorous and transparent data collection process (e.g., recording demographic and identity information of annotators) \cite{cheng2021socially} to help reduce downstream biases. Challenging sets to measure application bias such as \cite{rottger2021hatecheck} need to be created to test the robustness of debiasing methods.
\section*{Ethic Statement}
We heavily rely on existing bias and fairness metrics, which certainly have no guarantee of \textit{unbiased} word embeddings or machine learning models. In fact, most metrics can only be considered an indicator of bias at most \cite{seraphina2021intrinsic}, especially since significant limitations w.r.t. these metrics have been found \cite{garrido2021survey}. Therefore, we urge practitioners not to rely on these debiased word embeddings alone, but also at least consider bias mitigation in specific downstream tasks. Further, we did not discuss many other negative impacts of language models that practitioners should consider, such as high energy consumption or not including all stakeholders in the design phase \cite{bender2021dangers}. 
\section*{Acknowledgements}
This work is supported by the Office of Naval Research (ONR) under the grant N00014-21-1-4002 and National Science Foundation (NSF) under the grants \#2036127 and \#2114789. Lu Cheng is also supported by the startup fund provided by the Department of Computer Science at the University of Illinois Chicago. The views, opinions, and/or findings expressed are the authors’ and should not be interpreted as representing the official views or policies of the funding agency.
\bibliography{anthology}
\bibliographystyle{acl_natbib}
\appendix
\section{Additional Empirical Results for Validating the Intersectional Hypothesis}
\begin{figure}
    \centering
    \includegraphics[width=.65\linewidth]{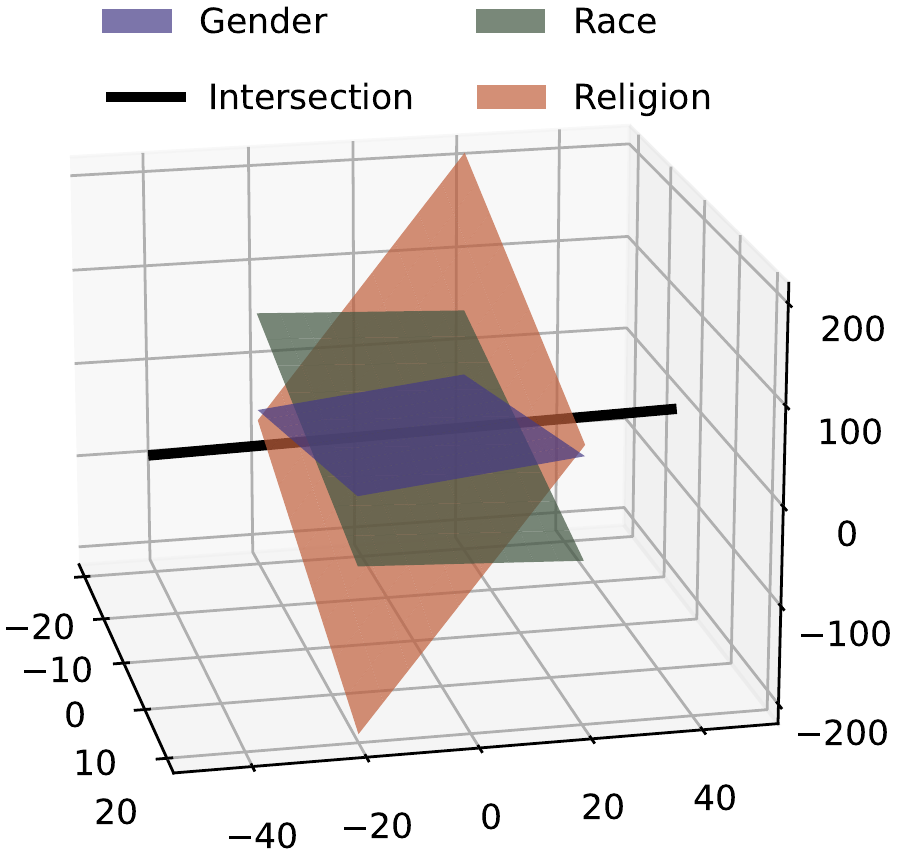}
   \caption{The geometry of Gender, Race, and Religion bias subspaces and the intersectional subspaces (Intersection).}
\label{fig:visualize3}
\end{figure}

Following the same procedure in Sec. 4.2.1, we draw subspaces w.r.t. all of three categories as 2-dimensional planes in Figure \ref{fig:visualize3}. We observe similar results to those w.r.t. two categories.
\section{Stimuli for Identifying Intersectional Groups and Biases}
The stimuli used to construct the ``ground-truth'' intersectional bias subspace are from \cite{caliskan2017semantics} and \cite{parada2016ethnolinguistic}. We follow \cite{guo2020detecting} to identify the names of intersectional groups and corresponding attributes associated with the intersectional biases. This process is further verified by human subjects \cite{ghavami2013intersectional}. We use this validation set to evaluate the quality of the intersectional subspace constructed by JoSEC. In particular, the ground-truth intersectional subspace is used in the empirical validation of the Intersectional Hypothesis (Appendix A) and comparisons of the debiasing performance between JoSEC and Hard\_Insec. 

Names to identify intersectional groups:
\begin{itemize}
    \item \textbf{African American females}: Aisha, Keisha, Lakisha, Latisha, Latoya, Malika, Nichelle, Shereen, Tamika, Tanisha, Yolanda, Yvette
    \item \textbf{African American males}: Alonzo, Alphonse, Hakim, Jamal, Jamel, Jerome, Leroy, Lionel, Marcellus, Terrence,
Tyrone, Wardell
\item \textbf{European American females}: Carrie, Colleen, Ellen,
Emily, Heather, Katie, Megan, Melanie, Nancy, Rachel,
Sarah, Stephanie
\item \textbf{European American males}: Andrew, Brad, Frank, Geoffrey, Jack, Jonathan, Josh, Matthew, Neil, Peter, Roger,
Stephen
\item \textbf{Mexican American females}: Adriana, Alejandra, Alma, Brenda, Carolina, Iliana, Karina, Liset, Maria, Mayra, Sonia, Yesenia
\item \textbf{Mexican American males}: Alberto, Alejandro, Alfredo,
Antonio, César, Jesús, José, Juan, Miguel, Pedro, Rigoberto, Rogelio
\end{itemize}

Attributes indicating intersectional biases:
\begin{itemize}
    \item \textbf{Intersectional biases of African American females}:  aggressive, athletic, bigbutt, confident, darkskinned, friedchicken, ghetto, loud, overweight, promiscuous, unfeminine, unintelligent, unrefined
     \item \textbf{Intersectional biases of African American males}:  athletic, criminals, dangerous, darkskinned, gangsters, hypersexual, lazy, loud, poor, rapper, tall, unintelligent, violent
     \item \textbf{Intersectional Biases of European American females}:
arrogant, attractive, blond, ditsy, emotional, feminine, highstatus, intelligent, materialistic, petite, racist, rich, submissive, tall
\item \textbf{Intersectional biases of European American males}: allAmerican, arrogant, attractive, blond, high-status, intelligent, leader, privileged, racist, rich, sexist, successful,
tall
\item \textbf{Intersectional biases of Mexican American females}: cook, curvy, darkskinned, feisty, hardworker, loud, maids,
promiscuous, sexy, short, uneducated, unintelligent
\item \textbf{Intersectional biases of Mexican American males}:   aggressive, arrogant, darkskinned, day-laborer, drunks, hardworker, illegal-immigrant, jealous, macho, poor, promiscuous, short, uneducated, unintelligent, violent
\end{itemize}
\section{Additional Experimental Results for RQ. 3}
This section presents results for the semantic utility of debiased word embeddings under the \textit{Model Retraining} paradigm. In particular, we show Precision, Recall, F1 scores w.r.t. NER Tagging, POS Tagging, and POS Chunking, respectively. We retrain the model with the debiased word embeddings. As shown in Table \ref{downstream2}, we observe similar results to those using Embedding Replacement. Together with the results in Sec. 5.3.3, we show that debiasing for multiple social categories using the hard-debiasing method does not have a significant influence on the semantic utility of word embeddings. This applies to both joint and intersectional debiasing.
\label{sec:appendix}
\begin{table*}[t]
\centering
\caption{Utility of biased and debiased word embeddings in NER Tagging (NER), POS Tagging (POS-T), and POS Chunking (POS-C) tasks, under the \textit{Model Retraining} paradigm. Word embeddings are debiased by hard-debiasing with different subspaces. $\Delta$ denotes the change before and after debiasing.}
\resizebox{\textwidth}{!}{%
\begin{tabular}{|c|c|c|c|c|c|c|c|c|c|c|c|c|c|c|c|}
\hline
                   & \multicolumn{3}{c|}{Hard\_Seq} & \multicolumn{3}{c|}{Hard\_Insec} & \multicolumn{3}{c|}{SUM} & \multicolumn{3}{c|}{MEAN} & \multicolumn{3}{c|}{JoSEC} \\ \hline
Tasks              & NER      & POS-T    & POS-C    & NER       & POS-T     & POS-C    & NER    & POS-T  & POS-C  & NER     & POS-T  & POS-C  & NER     & POS-T   & POS-C  \\ \hline
Biased F1          & 0.97     & 1.00     & 0.99     & 0.97      & 1.00      & 0.99     & 0.97   & 1.00   & 0.99   & 0.97    & 1.00   & 0.99   & 0.97    & 1.00    & 0.99   \\ \hline
$\Delta$ F1        & -0.01    & 0.01     & 0.00     & -0.01     & 0.00      & 0.00     & -0.01  & 0.01   & 0.00   & -0.00   & 0.00   & 0.00   & -0.00   & 0.00    & 0.00   \\ \hline
$\Delta$ Precision & -0.00    & 0.00     & 0.00     & -0.01     & 0.00      & 0.00     & -0.01  & 0.00   & 0.00   & 0.00    & 0.00   & 0.00   & -0.00   & 0.00    & 0.00   \\ \hline
$\Delta$ Recall    & -0.02    & 0.03     & 0.02     & -0.02     & 0.01      & 0.02     & -0.01  & 0.02   & 0.01   & -0.01   & 0.01   & 0.01   & -0.00   & 0.01    & 0.01   \\ \hline
\end{tabular}%
}
\label{downstream2}
\end{table*}
\section{Experimental Setup for Toxicity Classification in RQ. 4}
We employ LSTM (Long Short-Term Memory) \cite{hochreiter1997long} as the toxicity classifier, which takes the input of a variety of word embeddings considered in this work. We then measure the biases using two commonly adopted metrics \cite{dixon2018measuring}: False Positive Equality Difference (FPED) and False Negative Equality Difference (FNED). 
FNED/FPED is defined as the sum of deviations of group-specific False Negative Rates (FNRs)/False Positive Rates (FPRs) from the overall FNR/FPR.
Given $N$ demographic groups (e.g., female and male in gender) and we denote each group as $G_{i \in \{1,...,N\}}$, FNED and FPED are calculated as:
\begin{equation}
\begin{aligned}
    FNED&=\sum_{i \in \{1,...,N\}} |FNR-FNR_{G_i}|,\\
    FPED&=\sum_{i \in \{1,...,N\}} |FPR-FPR_{G_i}|.
\end{aligned}
\end{equation}
\noindent where $FNR_{G_i}$ denotes the FNR calculated over group $G_i$ and FNR is calculated over the entire training set. A debiased model is expected to have similar FNR and FPR for different groups belonging to the same identity, therefore, smaller FNED and FPED are desired.
Ideally, the sum of FNED and FPED is close to zero.
\end{document}